\documentclass[sigconf]{acmart}
\usepackage{amsmath}
\usepackage{booktabs}
\usepackage{multirow}
\usepackage[table]{xcolor}
\usepackage[most]{tcolorbox}
\usepackage{pifont}
\setcopyright{acmlicensed}
\copyrightyear{2025}
\acmYear{2025}
\acmDOI{10.1145/3746252.3761340}
\acmConference[CIKM '25] {Proceedings of the 34th ACM International Conference on Information and Knowledge Management}{ November 10--14, 2025}{Seoul, Republic of Korea.}
\acmBooktitle{Proceedings of the 34th ACM International Conference on Information and Knowledge Management (CIKM '25), November 10--14, 2025, Seoul, Republic of Korea}
\acmISBN{979-8-4007-2040-6/2025/11}

\begin{document}

\title{Multi-Source Knowledge Pruning for Retrieval-Augmented Generation: A Benchmark and Empirical Study}


\author{Shuo Yu}
\orcid{0009-0006-1060-5451}
\affiliation{%
  \institution{State Key Laboratory of Cognitive Intelligence, University of Science and Technology of China}
  \city{Hefei}
  \state{}
  \country{China}
}
\email{yu12345@mail.ustc.edu.cn}

\author{Mingyue Cheng}
\authornote{Mingyue Cheng is the corresponding author.}
\orcid{0000-0001-9873-7681}
\affiliation{%
  \institution{State Key Laboratory of Cognitive Intelligence, University of Science and Technology of China}
  \city{Hefei}
  \state{}
  \country{China}
}
\email{mycheng@ustc.edu.cn}

\author{Qi Liu}
\orcid{0000-0001-6956-5550}
\affiliation{%
  \institution{State Key Laboratory of Cognitive Intelligence, University of Science and Technology of China}
  \city{Hefei}
  \state{}
  \country{China}
}
\email{qiliuql@ustc.edu.cn}

\author{Daoyu Wang}
\orcid{0009-0002-0452-0516}
\affiliation{%
  \institution{State Key Laboratory of Cognitive Intelligence, University of Science and Technology of China}
  \city{Hefei}
  \state{}
  \country{China}
}
\email{wdy030428@mail.ustc.edu.cn}

\author{Jiqian Yang}
\orcid{0009-0007-6421-2626}
\affiliation{%
  \institution{State Key Laboratory of Cognitive Intelligence, University of Science and Technology of China}
  \city{Hefei}
  \state{}
  \country{China}
}
\email{yangjq@mail.ustc.edu.cn}

\author{Jie Ouyang}
\orcid{0009-0001-7652-368X}
\affiliation{%
  \institution{State Key Laboratory of Cognitive Intelligence, University of Science and Technology of China}
  \city{Hefei}
  \state{}
  \country{China}
}
\email{ouyang\_jie@mail.ustc.edu.cn}

\author{Yucong Luo}
\orcid{0000-0003-0685-0834}
\affiliation{%
  \institution{State Key Laboratory of Cognitive Intelligence, University of Science and Technology of China}
  \city{Hefei}
  \state{}
  \country{China}
}
\email{prime666@mail.ustc.edu.cn}

\author{Chenyi Lei}
\orcid{0000-0001-6287-3673}
\affiliation{%
 \institution{Kuaishou Technology}
  \city{Beijing}
  \state{}
  \country{China}
}
\email{leichy@mail.ustc.edu.cn}

\author{Enhong Chen}
\orcid{0000-0002-4835-4102}
\affiliation{%
  \institution{State Key Laboratory of Cognitive Intelligence, University of Science and Technology of China}
  \city{Hefei}
  \state{}
  \country{China}
}
\email{cheneh@ustc.edu.cn}




\renewcommand{\shortauthors}{Shuo Yu et al.}

\begin{abstract}
Retrieval-augmented generation (RAG) is increasingly recognized as an effective approach to mitigating the hallucination of large language models (LLMs) through the integration of external knowledge.  While numerous efforts, most studies focus on  a single type of external knowledge source.  However, in real-world applications, most situations involve diverse  knowledge from various sources, yet this area has been less explored. The main dilemma is the lack of a suitable dataset containing multiple knowledge sources and pre-exploration of the associated  issues. To address these challenges, we standardize a benchmark dataset that combines structured and unstructured knowledge across diverse and complementary  domains. Based on this dataset, we further develop a plug-and-play RAG framework, \textbf{PruningRAG}, whose main characteristic is the use of multi-granularity pruning strategies to optimize the integration of relevant information while minimizing misleading context. It consistently improves performance across various existing RAG variants, demonstrating its robustness and broad applicability.  Building upon the standardized dataset and PruningRAG, we also report  a series of  experimental results, as well as insightful findings. Our dataset and  code  are publicly available\footnote{https://github.com/USTCAGI/PruningRAG}, with the aim of advancing future research in the RAG community.

\end{abstract}






\maketitle

\section{Introduction}

\begin{figure}[t]
    \centering

    \includegraphics[width=1\linewidth]{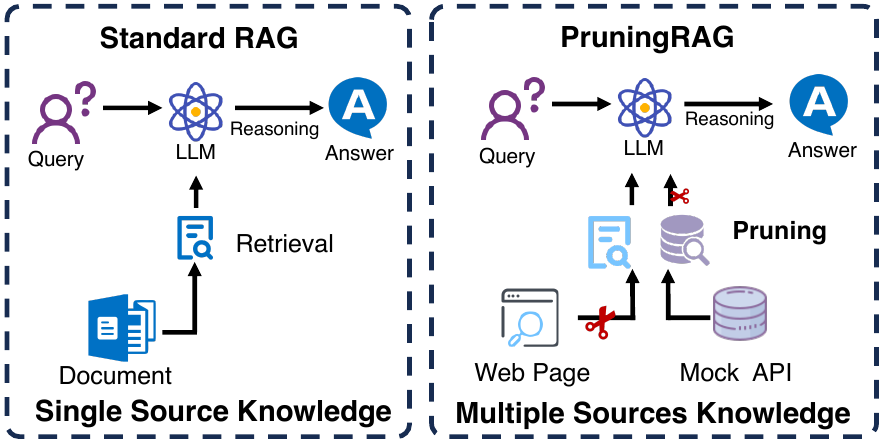}

    \caption{Comparison between Standard RAG and PruningRAG. Standard RAG typically relies on a single knowledge source for retrieval and generation. In contrast, PruningRAG enhances the utilization of multiple external knowledge sources by applying multi-granularity pruning strategies. }

    \label{fig:enter-label}
\end{figure}
In recent years, the excellent performance of large language models (LLMs)~\cite{brown2020language,jiang2023reformulating,luo2023unlocking,rag_survey} across various tasks in diverse domains has attracted widespread attention from researchers. However, despite their continuous improvement, LLMs rely solely on internal knowledge stored in their parameters during pretraining, which is inherently limited and cannot be updated in real time. Consequently, they are often susceptible to hallucinations, particularly when handling long-tail or time-sensitive queries~\cite{zhou2020detecting,10.1145/3571730}. To address this dilemma, retrieval-augmented generation (RAG)~\cite{lewis2020retrieval, guu2020retrieval, cheng2024general}  integrates retrieved context as external knowledge to enhance the capabilities of LLMs.
It serves as a bridge between the static and often limited internal knowledge of LLMs and the extensive, dynamically evolving information in the real world, thereby mitigating hallucinations and enhancing reliability.

Numerous studies on RAG have been proposed to effectively integrate external knowledge with the internal knowledge of LLMs~\cite{hyde, Chan2024RQRAGLT, su2024dragin}. Through a review of current research on RAG, we found that most studies primarily focus on the utilization of a single knowledge source. However, practical applications often require access to multiple knowledge sources, which can vary significantly in format, timeliness, and domain. This necessitates intelligent selection of knowledge sources based on the characteristics of each query, as well as the design of retrieval strategies tailored to the specific properties of the selected sources.
Despite this need, research on RAG with multiple external knowledge sources remains limited, primarily due to the absence of standardized benchmark datasets and a lack of investigation into how diverse knowledge sources interact and affect model performance.


Fortunately, we found that the KDD Cup 2024 CRAG competition dataset~\cite{yang2024crag,kddcup} comprises two distinct types of external knowledge sources: unstructured web pages of variable quality with limited timeliness but broad coverage, and APIs, which offer structured accurate information with strong real-time performance\cite{table_survey,oneperson,personeval}. 
However, the dataset still encounters some challenges in its suitability for broad research applications. For instance, the HTML-formatted web page data it contains present significant challenges for LLM processing, with no unified standards currently available for cleaning and parsing these data. Furthermore, how to effectively prune multi-source external knowledge and reduce misleading information has been less explored.
In this paper, we standardize the dataset and establish a new benchmark, providing a solid foundation for future research in the field. 
To standardize this dataset, we undertake significant efforts. For example, we clean the web page knowledge by removing excessive HTML tags and converting it into an LLM-friendly Markdown format. Additionally, we use a rule-based approach to convert API responses into natural language, enhancing data quality, ensuring compatibility with existing RAG frameworks, and enabling fair evaluation.

Building upon this dataset, we introduce PruningRAG, a new framework for RAG that performs multi-granularity pruning of diverse knowledge sources. Coarse-grained pruning effectively removes misleading information from inappropriate sources, thereby mitigating hallucinations. Meanwhile, adaptive fine-grained pruning further refines the knowledge retained from coarse-grained pruning, ensuring higher relevance while reducing extraneous information and thereby improving overall accuracy. For web-based knowledge, a two-stage retrieval strategy is employed, combining sparse and dense retrieval to enhance both coverage and precision. In contrast, for API-based knowledge, named entity recognition (NER) is first applied to the query to extract key entities, which are then used to route the query to the most appropriate API endpoint.  After obtaining pruned knowledge, we use strategies such as chain-of-thought (CoT)~\cite{cot} and in-context learning (ICL)~\cite{dong2022survey} to enhance the performance of reasoning. In addition, our framework is plug-and-play, enabling integration with a variety of existing RAG variants with minimal modification.

Based on our dataset and framework, we conduct extensive experiments and report key insights. For coarse-grained pruning, a fine-tuned LLM can dynamically select relevant knowledge sources to maximize utility while minimizing misleading context. In contrast, an untuned LLM may over-prune, leading to harmful outcomes. In fine-grained pruning, sparse retrieval can serve as a pre-filtering step to narrow the retrieval scope and improve efficiency. Within this reduced candidate set, dense retrieval outperforms both sparse and hybrid approaches. Notably, we observe that the effectiveness of CoT prompting varies depending on the type of knowledge source. In addition, both the domain alignment between the few-shot examples and the query, as well as the relative positioning of the query and the retrieved context within the prompt, significantly influence reasoning performance.

In summary, our paper makes several key contributions:

\begin{itemize}
    \item We standardize a benchmark dataset that integrates structured and unstructured external knowledge across diverse and complementary domains .
    
    \item We develop PruningRAG, a plug-and-play framework that employs a multi-granularity pruning strategy  to optimize the integration and utilization of multi-source knowledge.
    
    \item  We conduct extensive experiments and present key insights into leveraging multi-source knowledge, aiming to inform and support future research.

\end{itemize}

\section{Related Work}

\subsection{Retrieval-Augmented Generation}

RAG~\cite{lewis2020retrieval} has emerged as a strong approach to mitigating hallucinations in LLMs by incorporating external knowledge. Early methods utilized a straightforward “retrieve-then-generate” pipeline, whereas more advanced frameworks now integrate query refinement~\cite{RRR,Chan2024RQRAGLT}, iterative retrieval~\cite{Iter-RetGen,self-ask}, and modular architectures~\cite{Gao2023RetrievalAugmentedGF}.
Query refinement improves the semantic understanding of the original question and better aligns the query with relevant documents, thereby enhancing retrieval precision. Iterative retrieval enables the model to extract useful information over multiple retrieval steps, which is particularly beneficial for complex multi-hop reasoning tasks. Modular architectures provide greater flexibility by decoupling system components, allowing targeted optimization across different dimensions such as retrieval accuracy, generation fluency, and context integration. Dynamic retrieval frameworks, such as Self-RAG~\cite{asai2024selfrag} and DRAGIN~\cite{su2024dragin}, progressively refine retrieved information based on intermediate signals, enabling more targeted and efficient retrieval by avoiding unnecessary queries.  GraphRAG~\cite{edge2024localglobalgraphrag} introduces a graph-based indexing mechanism that captures the global structure of document relationships, facilitating richer context modeling.

However, these advancements often focus solely on the efficient utilization of a single knowledge source, while generally overlooking the complexities involved in managing multiple  sources of knowledge. Although some methods incorporate multiple sources of knowledge~\cite{sarmah2024hybridrag,wang2024unims,zhao2024towards,hoh}, they often lack diversity in data formats, fields, or timeliness. To bridge this gap, we propose PruningRAG, which uses multi-granularity pruning to reduce misleading information, thereby integrating multi-source knowledge.

\subsection{Existing Benchmarks for RAG}

As RAG frameworks evolve, new benchmarks have emerged to measure and guide their capabilities. For instance, RGB~\cite{Chen2023BenchmarkingLL} evaluates robustness, integration, and counterfactual reasoning; CRUD-RAG~\cite{10.1145/3701228} follows a structured Create-Read-Update-Delete workflow; RAGBench~\cite{Friel2024RAGBenchEB} focuses on explainability with detailed metrics; and RAGEval~\cite{zhu2024ragevalscenariospecificrag} automates dataset generation for rigorous testing. These benchmarks offer a comprehensive framework for assessing RAG performance. In addition, ARES\cite{ares} and RAGAS\cite{ragas} are automated, reference-free evaluation tools. RAGAS focuses on metrics such as context relevance, answer faithfulness, and answer relevance, while ARES evaluates similar dimensions with an emphasis on scoring precision and statistical robustness.


However, most existing benchmarks focus on single-source knowledge integration and do not evaluate the utilization of multi-source knowledge.  Although the CRAG benchmark~\cite{yang2024crag} incorporates both web pages and API sources, its dataset lacks standardized HTML parsing and hinders LLMs from effectively utilizing JSON-formatted API information. To address these limitations, we standardize the dataset and introduce a new benchmark to tackle multi-source heterogeneous knowledge integration, reduce hallucinations, and enhance reasoning capabilities.

\section{Preliminaries}
This section introduces the formulation of retrieval-augmented generation (RAG) in the scenario of multi-source external knowledge. We then describe the standardization process of our benchmark dataset, followed by an overview of its characteristics and the unique challenges it poses to existing RAG systems.
\subsection{Problem Formulation}

We consider a query \( q \) and a set of external knowledge sources \(\mathcal{K} = \{ K_1, K_2, \dots, K_p \}\), where each source \(K_i\) is associated with a document set \(\mathcal{D}_{K_i}\). Let \(\mathcal{D} = \bigcup_{i=1}^p \mathcal{D}_{K_i}\) denote the complete collection of retrievable documents. The goal is to generate an answer \( A \) by selecting and merging documents most relevant to \( q \) and then using LLM to produce the final response. Formally, let \(\mathcal{D}_q\) be the subset of \(\mathcal{D}\) retrieved for \( q \). The final answer is given by:
\begin{equation}
\label{eq:toprank}
\operatorname{Answer} = \operatorname{LLM}\bigl(\mathcal{D}_q \oplus q\bigr),
\end{equation}
where \(\oplus\) indicates the concatenation of \(\mathcal{D}_q\) and \(q\).

\begin{table}[t]
\centering
\caption{An example from the standardized dataset, illustrating the unified input format of RAG. Each QA pair corresponds to 5 or 50 web pages, each of which contains three fields: page\_name, page\_snippet, and page\_result.}
\label{tab:crag-example}
\renewcommand{\arraystretch}{1.1}
\setlength{\tabcolsep}{8pt}
\resizebox{1.0\columnwidth}{!}{
\begin{tabular}{>{\raggedright\arraybackslash}m{0.25\linewidth} p{0.68\linewidth}}
\toprule
\textbf{Field} & \textbf{Value} \\
\midrule
interaction\_id & ... \\

domain & movie \\

question\_type & simple \\

static\_or\_dynamic & static \\
\midrule
query\_time & 03/19/2024, 23:22:04 PT \\
query & can you tell me the date that randall wallace was born? \\

answer & 1949-07-28 \\

\midrule
page\_name & Randall Wallace-Wikipedia \\

page\_snippet & Randall Wallace is an American screenwriter... \\

page\_result & Randall Wallace (born July 28, 1949) is an  American screenwriter, film director and ... \\
\bottomrule
\end{tabular}}

\end{table}

\begin{figure*}
	\centering
	\includegraphics[width=1\linewidth]{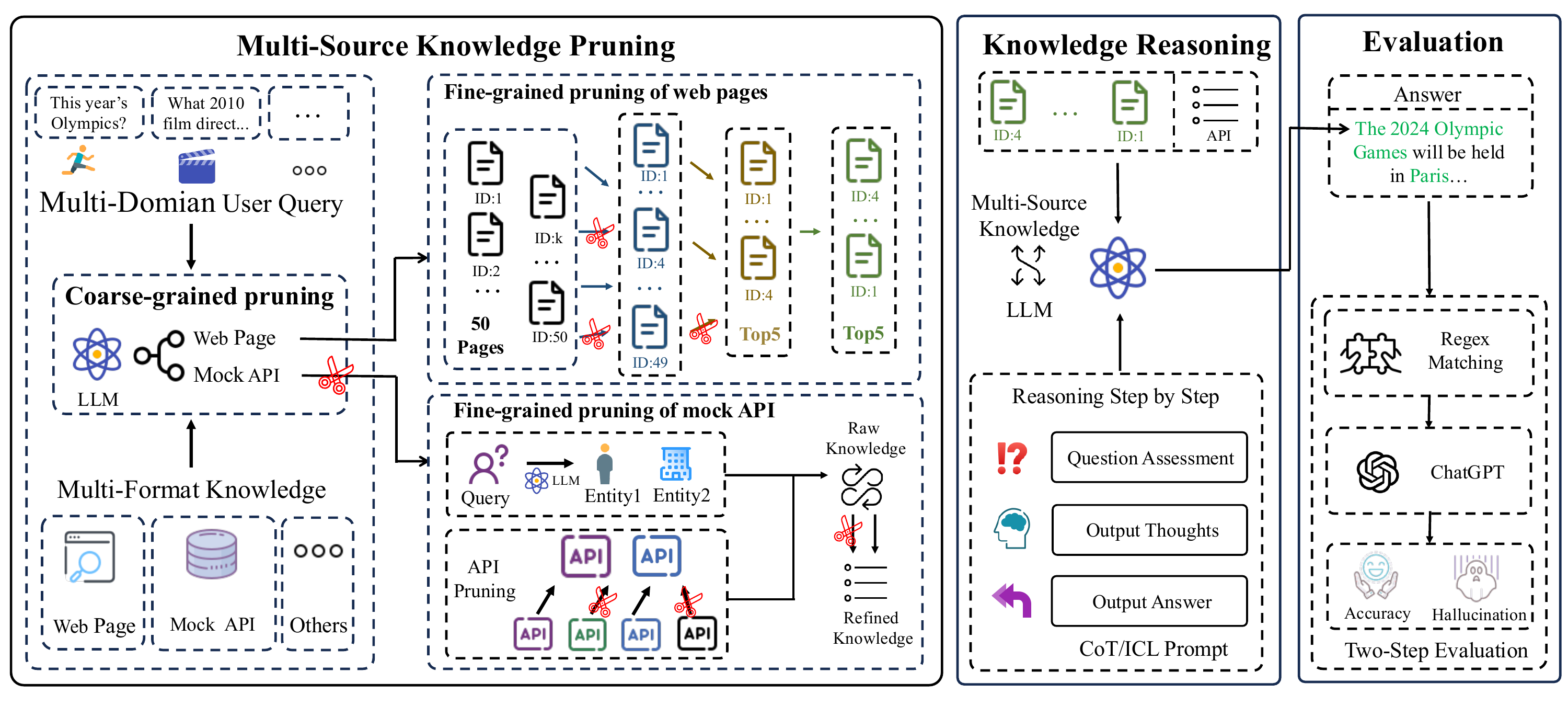}

	\caption{ An illustration of  PruningRAG, including multi-source knowledge pruning, knowledge reasoning and evaluation. Knowledge pruning filter out irrelevant knowledge sources and improve context relevance. The pruned knowledge is combined with the query to reason and the answer is evaluated based on accuracy and hallucination.}
\Description{}
	\label{fig:our-RAG}
\end{figure*}

\subsection{ A Multi-Source Knowledge RAG Dataset}
\subsubsection{Dataset Standardization}
While numerous datasets have been proposed to support retrieval-augmented generation (RAG)~\cite{kwiatkowski2019natural,  joshi2017triviaqa, stelmakh2022asqa, tang2024multihop}, most rely on a single external knowledge source. However, real-world scenarios often require access to heterogeneous sources that vary in format, domain, and temporal relevance. To address this gap, we build upon the dataset released in the KDD Cup 2024 CRAG competition, which uniquely provides both unstructured web content (in HTML format) and structured knowledge accessed via APIs. Despite its potential, the original CRAG dataset poses several practical hurdles: HTML documents contain noisy and inconsistent tags that hinder knowledge extraction, and there is no unified parsing standard across sources. Moreover, API responses contain a large amount of information formatted in JSON, which is not directly suitable for LLM-based reasoning.

To overcome these issues, we perform systematic preprocessing and standardization. For the web data, we clean and convert HTML pages into Markdown format, removing irrelevant tags and ensuring structural consistency for token-level processing. For the API data, we apply rule-based transformation by aligning entities with query terms and converting structured outputs into natural language statements, enabling seamless integration with LLM prompts. These refinements produce a clean and unified multi-source benchmark better suited for RAG research, where both retrieval effectiveness and reasoning performance can be fairly evaluated.

\subsubsection{Dataset Overview}
The resulting dataset contains 4,409 QA pairs spanning diverse domains such as finance, sports, and entertainment, and temporal categories including static and real-time queries. It supports eight distinct question types, such as simple factual questions, conditional logic, and multi-hop reasoning. The dataset is split into training, validation, and test subsets, with 1,371 QA pairs reserved for testing. Each query is associated with either five or fifty web pages, along with a structured API result derived from a knowledge graph comprising 2.6 million entities. To maintain consistency, we adopt a unified input format (illustrated in Table~\ref{tab:crag-example}), where only the query, timestamp (query\_time), and retrieved content are exposed to the model at inference time; metadata such as domain labels are used only for analysis. Notably, we do not construct a pre-indexed vector database for the web pages. Instead, we simulate realistic RAG scenarios by performing retrieval from the full corpus at query time, reflecting the immediacy of web-based knowledge access in practical applications.

Web pages are static, often noisy, and broad in coverage, making them better suited for open-domain factual queries. In contrast, API responses are precise and time-sensitive, though limited in scope, and are particularly effective for domains such as finance, where up-to-date information is critical for answering real-time queries. Some queries may lack support from either source, requiring the model to draw on its internal knowledge or abstain from answering when reliable information is unavailable, thereby mitigating the risk of hallucination.  Moreover, when irrelevant knowledge is retrieved particularly from the wrong source it can amplify hallucinations. These design characteristics reflect realistic complexities in multi-source information integration, setting our dataset apart from prior benchmarks that assume a single, answer-containing source.

\begin{figure}
    \centering
    \includegraphics[width=1\linewidth]{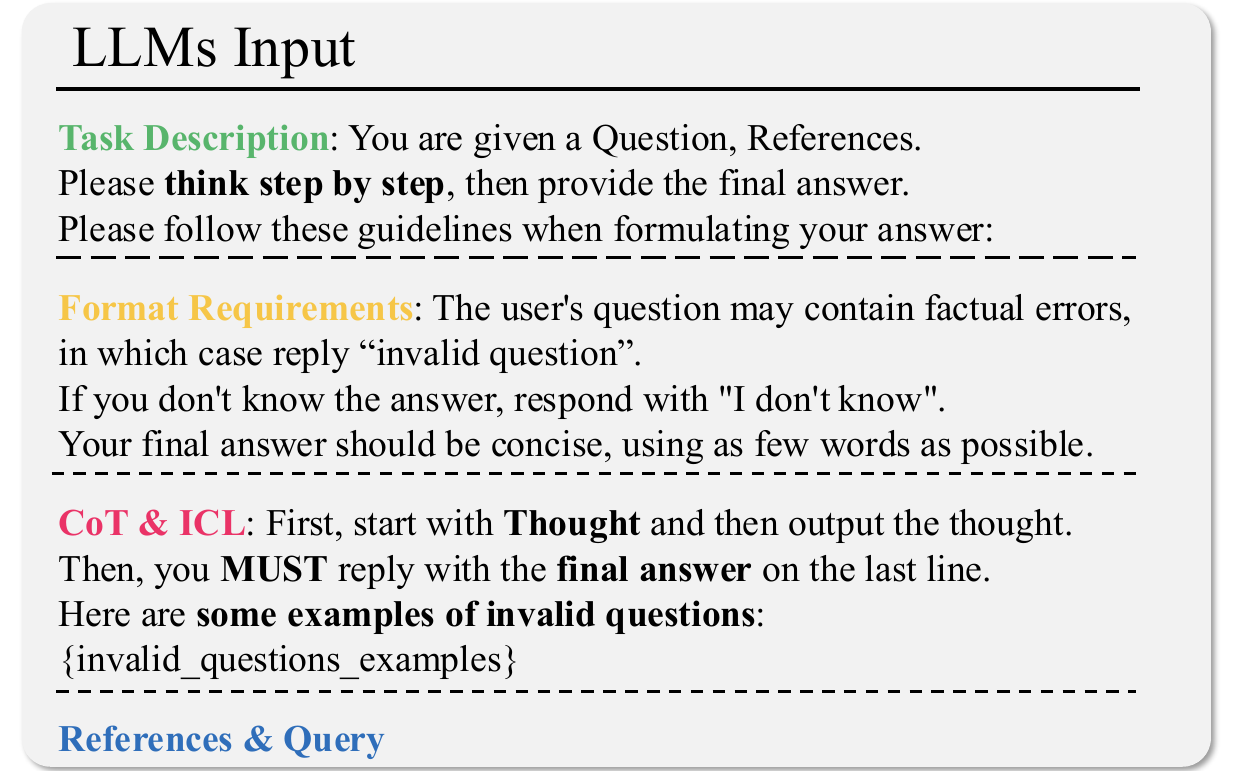}

    \caption{Prompt design template incorporating CoT and ICL  for enhanced reasoning.}

    \label{fig:prompt}
\Description{}
\end{figure}
\begin{table*}[ht]
\centering
\caption{Performance of RAG frameworks with and without PruningRAG on different knowledge sources.}
\begin{tabular}{c cc cc cc c c c cc c}
\toprule
& \multicolumn{2}{c}{Naive RAG} & \multicolumn{2}{c}{HyDE} & \multicolumn{2}{c}{Iter-RETGEN} & \multicolumn{2}{c}{RRR}  & \multicolumn{2}{c}{Self-Ask} & \multicolumn{2}{c}{Self-RAG} \\
\cmidrule(lr){2-3}\cmidrule(lr){4-5}\cmidrule(lr){6-7}
\cmidrule(lr){8-9}\cmidrule(lr){10-11}
\cmidrule(lr){12-13}
Knowledge Source & Acc. & Hall. & Acc. & Hall. & Acc. & Hall. & Acc. & Hall. & Acc. & Hall. & Acc. & Hall. \\
\cmidrule{1-13}
5 Web Pages         & 24.80 & 22.69 & 24.65 & 18.24 & 28.59 & 18.45 & 15.61 &  12.77 & 24.46 & 29.98 & 14.29 & 8.97 \\
50 Web Pages        & 32.82 & 29.54 & 30.48 & 17.06 & 31.50 & 25.16 & 19.76 &  15.31 & 30.78 & 26.26 & 14.51 & 8.75 \\
API                 & 32.38 & 11.67 & 32.75 & 11.46 & 32.38 & 11.67 & 31.66 & 12.48 & 25.23 & 17.80 & 14.36 & 8.82 \\
\cmidrule{1-13}
5 Web Pages + API   & 39.97 & 22.32 & 40.41 & 21.23 &  43.69 & 18.52 & 35.08 & 15.32 & 29.83 & 32.89 & 14.00 & 9.19 \\
\rowcolor{blue!15}
\quad + Pruning     & \textbf{44.56} & \textbf{21.23} & \textbf{43.03} & \textbf{18.31} & \textbf{43.83} & \textbf{17.21} & \textbf{36.03} &  \textbf{13.42} & \textbf{32.68} & \textbf{28.37} & \textbf{14.44} & \textbf{8.75} \\

\quad \(\uparrow\) Gain    & 11.48\% & 4.88\% & 6.48\% & 13.75\% & 0.32\% & 7.71\% & 2.71\% &  14.16\% & 9.55\% & 13.74\% & 3.14\% & 4.79\% \\
\cmidrule{1-13}
50 Web Pages + API  & 39.53 & 22.76 & 40.40 & 21.22 & 43.59 & 18.53 & 34.64 &  15.75 & 33.62 & 35.30 & 12.54 & 10.72 \\
\rowcolor{blue!15}
\quad + Pruning     & \textbf{43.15} & \textbf{18.01} & \textbf{41.76} & \textbf{15.97} & \textbf{43.85} & \textbf{17.21} & \textbf{35.77} & \textbf{14.58} & \textbf{33.84} & \textbf{32.45} & \textbf{13.03} & \textbf{9.34} \\

\quad \(\uparrow\) Gain    & 9.16\% & 20.87\% & 3.37\% & 24.74\% & 0.60\% & 6.58\% & 2.26\% &  7.43\% & 0.65\% & 8.07\% & 3.91\% & 12.87\% \\

\bottomrule
\end{tabular}

\label{table:rag_overall}
\end{table*} 

\section{Methodology}
In this section, we first present an overview of the entire framework, followed by the details of each component. The illustration of the PruningRAG framework is shown in Figure \ref{fig:our-RAG}.

\subsection{Overview of the PruningRAG Framework}


Our proposed framework, PruningRAG, is composed of three key components: multi-source knowledge pruning, knowledge reasoning, and performance evaluation. In the pruning stage, coarse-grained filtering first eliminates irrelevant or low-quality sources to effectively reduce the search space. This is followed by fine-grained pruning, which further refines the remaining candidates to ensure that only the most relevant contexts are preserved. The retained contexts are then integrated with the user query into a knowledge-enhanced prompt, where Chain-of-Thought (CoT) reasoning and In-Context Learning (ICL) are employed to guide the language model toward factually grounded outputs with reduced hallucination. Finally, a rigorous evaluation protocol—covering accuracy, hallucination rate, omission rate, and overall performance score—assessed through both string-matching and GPT-based evaluation methods, guarantees the reliability of the generated responses.

\subsection{Multi-Source Knowledge  Pruning}
In this section, we explain the specific strategies for pruning multi-source knowledge, including coarse-grained pruning to filter knowledge sources and fine-grained  pruning to obtain key context.
\subsubsection{ Coarse-Grained Knowledge  Pruning}

In scenarios involving multiple sources of external knowledge, relevant information may reside in 
\(\mathcal{K} = \{ K_1, K_2, \ldots, K_p \}\), within internal knowledge of LLM \(K_0\), or be completely unavailable. 
This makes pruning of irrelevant sources essential to prevent conflicts and hallucinations. We define a subset-selection 
function:
\begin{equation}
\label{eq:sourceselect}
\mathcal{K}_q = \mathrm{\Theta}\bigl(\{K_i\}_{i=0}^p, q\bigr),
\end{equation}
where \(\mathrm{\Theta}\) identifies the knowledge sources most appropriate for query \(q\). An LLM is used to
determine \(\mathcal{K}_q\). However,  initial experiments\footnote{The experimental results are detailed in Section~\ref{sec:coarse-pruning}.} showed that prompting the LLM with only \(q\) was insufficient for accurate source selection.

To address this issue, we constructed a specialized training dataset by annotating each query with the subset of knowledge sources that yielded correct answers during training. For queries where none of the available sources produced accurate responses, we instead selected sources that demonstrated the highest aggregate accuracy on similar queries. Using this dataset of (query, source subset) pairs, we fine-tuned the LLM to suppress low-relevance sources while preserving those essential for accurate answer generation.

\subsubsection{ Fine-Grained Knowledge Pruning}

For each selected knowledge source \(K_i \in \mathcal{K}_q\), we apply source-specific fine-grained pruning to extract relevant information for query \(q\). If \(K_i\) corresponds to a web-based source, where documents are stored in raw text form rather than as pre-encoded dense vectors, directly embedding all documents for dense retrieval would be computationally expensive.
To address this, we adopt a  hierarchical retrieval strategy.
We first apply sparse retrieval using BM25~\cite{cheng2021learning}:
\begin{equation}
\label{eq:sparse}
\mathcal{D}^{\mathrm{sparse}}_{K_i, q} = \mathrm{BM25}(\mathcal{D}_{K_i}, q),
\end{equation}
where \(\mathcal{D}^{\mathrm{sparse}}_{K_i, q}\) is the set of BM25-retrieved documents from \(\mathcal{D}_{K_i}\).

We then apply dense passage retrieval(DPR) to extract fine-grained, semantically relevant chunks:
\begin{equation}
\label{eq:dense}
\mathcal{D}_{K_i, q} = \mathrm{DPR}(\mathcal{D}^{\mathrm{sparse}}_{K_i, q}, q),
\end{equation}
where \(\mathcal{D}_{K_i, q}\) represents the set of high-relevance context.

If \(K_i\) is an API-based source, pruning involves both filtering out irrelevant APIs and narrowing the retrieved content. To this end, we first apply named entity recognition (NER) to extract key entities from the query, which guide the API in generating focused responses. We further employ query routing to map queries to appropriate APIs based on their functional scope. The structured outputs returned by the APIs are then converted into natural language through rule-based post-processing, resulting in a refined textual representation denoted by \(\mathcal{D}_{K_i, q}\).

Finally, the complete input context is constructed by aggregating the pruned content from all selected sources:
\begin{equation}
\label{eq:final_dq}
\mathcal{D}_q = \bigcup_{K_i \in \mathcal{K}_q} \mathcal{D}_{K_i, q},
\end{equation}
where \(\mathcal{D}_q\) serves as the final context input to the LLM.

\begin{table*}[ht]
\centering
\caption{Performance comparison of LLMs with varying parameter scales with and without PruningRAG.}
\begin{tabular}{c ccc ccc ccc ccc}
\toprule
Naive RAG & \multicolumn{3}{c}{Llama-3.1-70B-Inst.} & \multicolumn{3}{c}{Llama-3.1-8B-Inst.} & \multicolumn{3}{c}{Llama-3.2-3B-Inst.} & \multicolumn{3}{c}{Llama-3.2-1B-Inst.} \\
\cmidrule(lr){2-4} \cmidrule(lr){5-7} \cmidrule(lr){8-10} \cmidrule(lr){11-13}
Knowledge Source & Acc. & Hall. & Score & Acc. & Hall. & Score & Acc. & Hall. & Score & Acc. & Hall. & Score \\
\cmidrule{1-13}
5 Web Pages         & 30.49 & 15.31 & 15.18 & 24.80 & 22.69 & 2.11 & 29.68 & 23.63 & 6.05 & 12.83 & 16.33 & -3.50 \\
50 Web Pages        & 33.99 & 21.01 & 12.98 & 32.82 & 29.54 & 3.28 & 31.50 & 25.74 & 5.76 & 13.05 & 16.11 & -3.06 \\
API                 & 34.28 & 5.83  & 28.45 & 32.38 & 11.67 & 20.71 & 28.74 & 11.23 & 17.51 & 4.23  & 3.79  & 0.44 \\
\cmidrule{1-13}
5 Web Pages + API   & 47.84 & 18.16 & 29.68 & 39.97 & 22.32 & 17.65 & 36.90 & 25.02 & 11.88 & 16.19 & 16.92 & -0.73 \\
\rowcolor{blue!15}
\quad + Pruning     & \textbf{48.14} & \textbf{16.49} & \textbf{31.65} & \textbf{44.56} & \textbf{21.23} & \textbf{23.33} & \textbf{37.41} & \textbf{20.49} & \textbf{16.92} & \textbf{16.35} & \textbf{15.94} & \textbf{0.41} \\
\quad \(\uparrow\) Gain & 6.27\% & 9.20\% & 1.52\% & 11.48\% & 4.88\% & 4.83\% & 1.38\% & 18.11\% & 4.24\% & 0.99\% & 5.79\% & 1.15\% \\
\cmidrule{1-13}
50 Web Pages + API  & 47.99 & 19.54 & 28.45 & 39.53 & 22.76 & 16.77 & 38.41 & 24.00 & 14.41 & 15.17 & 19.62 & -4.45 \\
\rowcolor{blue!15}
\quad + Pruning     & \textbf{52.58} & \textbf{17.87} & \textbf{34.71} & \textbf{43.15} & \textbf{18.01} & \textbf{25.14} & \textbf{38.88} & \textbf{21.73} & \textbf{17.15} & \textbf{15.82} & \textbf{18.24} & \textbf{-2.42} \\
\quad \(\uparrow\) Gain & 9.56\% & 8.55\% & 4.87\% & 9.16\% & 20.87\% & 7.17\% & 1.22\% & 9.46\% & 2.39\% & 4.28\% & 7.03\% & 2.12\% \\
\bottomrule
\end{tabular}
\label{table:backbone}
\end{table*}

\subsection{Knowledge-Enhanced Reasoning}

As shown in Figure~\ref{fig:prompt}, we design a prompt that integrates CoT and ICL to optimize the use of pruned knowledge for reasoning. The prompt begins by clearly stating the task and instructing the model to answer based on the provided context. To mitigate hallucinations, the model is explicitly instructed to respond with “I don’t know” when the retrieved context is insufficient to support a confident and accurate answer.

To enhance reasoning capabilities, we include few-shot examples drawn from domains different from that of the query, which helps reduce overfitting to domain-specific patterns. The pruned knowledge is then presented, followed by the query and a CoT instruction that guides the model to reason step by step maximizing its reasoning potential while mitigating hallucinations.

Notably, we place the query after the retrieved content rather than before to avoid the lost-in-the-middle phenomenon in long prompts~\cite{liu2024lost}, thereby maintaining the model’s focus on the query. The prompt concludes by requesting both the reasoning process and a final, well-considered answer to ensure clarity and reliability.

\subsection{ Performance Evaluation}
We evaluate the performance of the RAG framework using four key metrics: accuracy (Acc.), hallucination (Hall.), missing (Miss.), and an overall score, defined as the difference between accuracy and hallucination. This score captures the model’s ability to extract relevant knowledge while minimizing misleading content, which can reflect the reliability of RAG.

The evaluation combines exact string matching with semantic assessments using GPT-based models. Specifically, a prediction is labeled as accurate if it exactly matches the ground truth. If the model outputs "I don't know," the response is categorized as missing. For all other cases, GPT-3.5 Turbo~\cite{ouyang2022training} is employed to perform a semantic comparison. If the predicted answer is judged to be semantically consistent with the ground truth, it is marked as accurate; otherwise, it is classified as a hallucination.

\section{Benchmark Evaluation of RAG}
In this section, we introduce the evaluation of PruningRAG and baselines in different knowledge sources using the standardized dataset, including experimental setup and analysis of the results.

\subsection{Experimental Setup}

\paragraph{Implementation details.}
For coarse-grained pruning, we adopt a fine-tuned Llama-3.1-8B-Instruct~\cite{dubey2024llama3herdmodels} to filter inappropriate knowledge sources. Fine-grained pruning combines BM25 for sparse retrieval with the BGE-M3 encoder~\cite{chen2024m3} for dense retrieval. Unless otherwise specified, Llama-3.1-8B-Instruct serves as the generator.

\begin{table}[t]
\centering
\caption{Comparison of performance of different strategies for leveraging knowledge sources. }
\label{Table:knowledge_selection}
\renewcommand\arraystretch{1.05}
\setlength{\tabcolsep}{7pt}
\begin{tabular}{lccc}
\toprule
\textbf{Experiment Setting} & \textbf{Acc.} & \textbf{Hall.} & \textbf{Score} \\
\midrule
LLM & 17.94 & 18.30 & -0.36 \\
Web Pages & 24.80 & 22.69 & 2.11 \\
API & 32.38 & 11.67 & 20.71 \\
Web Pages+API & 39.97 & 22.32 & 17.65 \\
\midrule
LLM+Web Pages & 17.94 & \textbf{10.14} & 7.80 \\
LLM+API & 40.55 & 18.30 & 22.25 \\
LLM+Web Pages+API & \textbf{45.73} & 31.36 & 14.37 \\
\midrule
LLM$\rightarrow$Web Pages & 25.30 & 31.14 & -5.84 \\
LLM$\rightarrow$API & 35.01 & 23.70 & 11.31 \\
LLM$\rightarrow$Web Pages+API & 38.22 & 31.58 & 6.64 \\
\midrule
Knowledge Source Pruning & 44.56 & 21.23 & \textbf{23.33} \\
\bottomrule
\end{tabular}

\end{table}

\begin{figure*}[t]
	\centering
	\includegraphics[width=\textwidth]{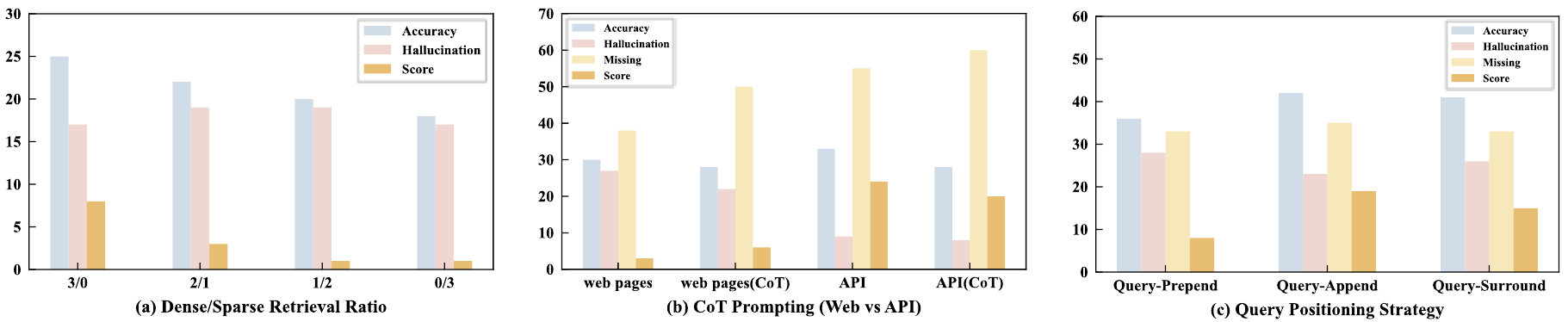}
	\caption{
Comparative analysis of retrieval methods and prompt strategies. 
(a) Performance comparison of sparse, dense, and hybrid retrieval. Hybrid retrieval combines top-ranked chunks from both sparse and dense retrieval, which are run in parallel, with selection based on weighted proportion.
(b) Comparison of Chain-of-Thought (CoT) prompting across different knowledge sources (web pages and API).
(c) Evaluation of three query positioning strategies within the prompt: 
Query-Prepend (placing the query before the retrieved context), 
Query-Append (placing the query after the context), and 
Query-Surround (placing the query both before and after the context).
}
	\label{fig:figure_5_6}
    \Description{}
\end{figure*}

\paragraph{Baselines.}

We apply our PruningRAG to a naive RAG baseline and five state-of-the-art frameworks: HyDE\cite{hyde}, Iter-RETGEN\cite{Iter-RetGen}, RRR\cite{RRR}, Self-Ask\cite{self-ask}, and Self-RAG\cite{asai2024selfrag}. The first four are train-free methods leveraging hypothetical document generation, iterative refinement, query reformulation, or question decomposition, respectively, while Self-RAG integrates retrieval and generation through fine-tuned self-refinement. All methods share identical dataset processing and evaluation protocols to ensure fair comparison and robust validation of our pruning strategy.

\subsection{Experimental Results}
Table~\ref{table:rag_overall} compares six RAG frameworks across three knowledge configurations—5 web pages, 50 web pages (which include the 5-page subset), and API—and evaluates the impact of integrating PruningRAG in terms of accuracy (Acc.) and hallucination rate (Hall.).
Across all configurations and methods, PruningRAG consistently improves performance by increasing accuracy and reducing hallucinations, demonstrating its general applicability. Notably, the reduction in hallucination is more significant than the gain in accuracy, suggesting that pruning plays a key role in mitigating misleading information, especially when retrieved content is noisy or only partially relevant. Among all the frameworks, Iter-RETGEN exhibits a relatively modest improvement in accuracy. This may be attributed to its iterative retrieval mechanism, which already ensures a high degree of knowledge utilization, thereby limiting the potential performance gain from pruning. Nevertheless, even in this scenario, hallucination rates are reduced, indicating that PruningRAG continues to enhance overall answer reliability by filtering misleading content.

\begin{table}[t]
\centering
\caption{Performance comparison across different model backbones and fine-tuning settings, with and without knowledge pruning.}
\label{tab:pruning_comparison}
\renewcommand\arraystretch{1.05}
\setlength{\tabcolsep}{7pt}
\begin{tabular}{c l l c c}
\toprule
\textbf{Prune} & \textbf{Model} & \textbf{Fine-tuning} & \textbf{Acc.} & \textbf{Score} \\
\midrule
\ding{55} & None &  & 39.97 & 16.65 \\
\midrule
\ding{51} & BERT-base & fine-tuned & 40.96 & 17.58 \\
\ding{51} & LLaMA-3.2-3B & frozen & 36.54 & 12.18 \\
\ding{51} & LLaMA-3.2-3B & fine-tuned & 43.25 & 18.16 \\
\ding{51} & LLaMA-3.1-8B & frozen & 36.11 & 22.17 \\
\ding{51} & LLaMA-3.1-8B & fine-tuned & \textbf{44.56} & \textbf{23.33} \\
\bottomrule
\end{tabular}

\end{table}

Table~\ref{table:backbone} presents a comparative analysis of LLMs with varying parameter scales within the naive RAG framework, along with the performance gains introduced by PruningRAG. Across all model sizes, PruningRAG consistently enhances accuracy and reduces hallucination, demonstrating its general effectiveness. Notably, the improvements are more pronounced for larger models (e.g., 70B and 8B variants), suggesting that models with greater capacity are better positioned to leverage the benefits of multi-granularity knowledge pruning. In contrast, the improvements observed in smaller models (e.g., 3B and 1B) are more modest, possibly due to their weaker semantic understanding capabilities and a greater reliance on redundant knowledge to support comprehension, which in turn limits the effectiveness of knowledge pruning. Nonetheless, even these smaller models experience reduced hallucination, highlighting the robustness of PruningRAG in improving knowledge reliability across model scales.

\section{Extensive Empirical Studies}

In this section, we leverage PruningRAG to conduct further experimental exploration on our dataset and present key insights from three perspectives: coarse-grained pruning, fine-grained pruning, and knowledge-enhanced reasoning.

\subsection{Impact of Coarse-Grained Pruning }
\label{sec:coarse-pruning}
Table \ref{Table:knowledge_selection} presents an evaluation of four knowledge utilization strategies. One approach relies exclusively on either the LLM’s internal knowledge or external knowledge. Another combines the LLM’s internal knowledge with one or more external sources to generate responses collaboratively. A further strategy prioritizes internal knowledge, consulting external sources only when the internal context is insufficient to produce an answer. Finally, our proposed method incorporates a knowledge source pruning mechanism to optimize the selection and integration of relevant knowledge.

The experimental results indicate that directly relying on multiple knowledge sources simultaneously often introduces conflicting information, resulting in performance degradation compared to using a single source. Additionally, prioritizing the internal knowledge of LLM before retrieval tends to generate hallucinations due to the inherent inaccuracies in the model’s internal knowledge. In contrast, our knowledge source pruning strategy dynamically  prunes knowledge sources based on the  characteristics of each query,  enabling the effective utilization of each knowledge source.

Table~\ref{tab:pruning_comparison} compares the effects of using different models for knowledge pruning. The results show that fine-tuning leads to notable improvements in accuracy, with the 8B-parameter model achieving the highest performance. This suggests that larger models possess stronger semantic understanding capabilities, resulting in more effective knowledge source pruning.

\begin{table}[t]
	\centering
	\caption{Comparison  of effectiveness and efficiency with and without sparse retrieval.}

		\begin{tabular}{c c c c c c}
			\toprule
			\textbf{Setting} & \textbf{Acc.} & \textbf{Hall.}  & \textbf{Latency(s)} \\ 
			\midrule
			w/ Sparse Retrieval    & \textbf{32.93} & 30.18  & \textbf{3.29} \\ 
			w/o Sparse Retrieval    & 32.82 & \textbf{29.54}  &  33.54 \\ 
			\bottomrule
		\end{tabular}

	\label{Table:pre-retrieval}
\end{table}

\begin{table}[t]
\centering
\caption{
Impact of few-shot example domain alignment on LLM reasoning. 
N* indicate that the few-shot examples are from a different domain than the query. 
}
\label{Table:few_shot}
\renewcommand\arraystretch{1.05}
\setlength{\tabcolsep}{6pt}
\begin{tabular}{ccccc}
\toprule
\textbf{Shots (N)} & \textbf{Acc.} & \textbf{Hall.} & \textbf{Miss.} & \textbf{Score} \\
\midrule
0     & 13.20 & \textbf{10.50} & 76.29 & 2.70 \\
\midrule
1     & 16.05 & 12.62 & 71.33 & 3.43 \\
2     & 16.12 & 12.98 & 70.90 & 3.14 \\
3     & 15.17 & 12.69 & 72.14 & 2.48 \\
\midrule
1*    & 16.12 & 11.89 & 71.99 & 4.23 \\
2*    & \textbf{18.02} & 11.23 & \textbf{70.75} & \textbf{6.78} \\
3*    & 16.41 & 11.60 & 72.00 & 4.81 \\
\bottomrule
\end{tabular}

\end{table}

\begin{figure*}[t]
	\centering
	\includegraphics[width=1.0\textwidth]{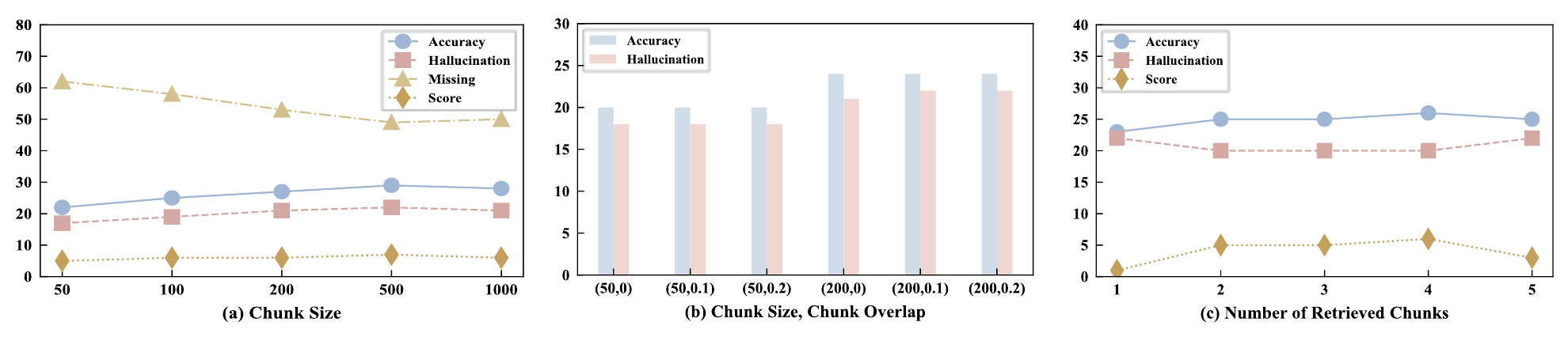}
	\caption{Sensitivity analysis of PruningRAG with respect to key retrieval configuration parameters.
\textbf{(a)} Effect of varying chunk size.
\textbf{(b)} Joint impact of chunk size and chunk overlap.
\textbf{(c)} Influence of the number of retrieved chunks.}
	\label{fig:figure_7_9}
\end{figure*}

\subsection{ Impact of Fine-Grained Pruning}

Table~\ref{Table:pre-retrieval} evaluates the necessity of incorporating an initial sparse retrieval step prior to fine-grained pruning. The results show that this sparse retrieval stage significantly reduces latency, especially when handling large volumes of external knowledge. By serving as a lightweight pre-filter, sparse retrieval narrows the candidate pool, thus enabling the subsequent dense retrieval to operate more efficiently and with lower computational cost. This design choice ensures context relevance while substantially improving system responsiveness in large-scale retrieval scenarios.

Figure~\ref{fig:figure_5_6} (a) further investigates the effectiveness of different fine-grained retrieval strategies after the initial sparse filtering. The results indicate that dense retrieval outperforms sparse retrieval in capturing semantic relevance. Moreover, combining dense and sparse retrieval (i.e., hybrid retrieval) leads to improved accuracy compared to sparse retrieval alone. However, this hybrid approach also introduces more hallucinations, suggesting that while it preserves more relevant information, it is less effective at filtering misleading context~\cite{cheng2022towards,gu2018search}. These findings highlight a trade-off between retrieval accuracy and hallucination control in multi-stage retrieval pipelines.

\subsection{Analysis of Knowledge Reasoning}

This section evaluates strategies to enhance LLM reasoning over pruned knowledge, including CoT prompting, few-shot examples, and adjusting query-context positioning. We also introduce a rejection mechanism based on confidence and context relevance to reduce hallucinations and improve reliability.
\subsubsection{Role of CoT reasoning.}

Figure \ref{fig:figure_5_6} (b) illustrates the varying impact of incorporating chain-of-thought (CoT) reasoning \cite{cot, trivedi2022interleaving} on performance, depending on the type of external knowledge sources. When internal LLM knowledge is combined with unstructured network data, which is often noisy and sparsely populated with relevant information, CoT’s step-by-step reasoning helps filter out irrelevant details and reduce hallucinations, thereby improving score. 
In contrast, when using APIs as external knowledge sources—where information is relatively precise—the hallucination rate of the LLM is already low, and the multi-step reasoning in CoT may result in overly cautious responses.
While this cautious approach reduces hallucinations, it may significantly compromise accuracy, even when the API provides reliable information.

\subsubsection{Impact of in-context learning.}
Table~\ref{Table:few_shot} presents the impact of incorporating few-shot examples on PruningRAG's performance. To assess generalization and reasoning, we compare two settings: one using domain-aligned examples and the other using cross-domain examples. Across all configurations, few-shot examples consistently improve accuracy over the zero-shot baseline, suggesting enhanced task understanding by the LLM.   Notably, cross-domain examples yield better results than domain-aligned ones, possibly due to reduced overfitting and increased variability, which enhance generalization and reasoning robustness. However, we also observe that increasing the number of few-shot examples does not always lead to better performance; excessive examples may introduce noise or encourage overfitting, suggesting that the number of demonstrations should be carefully balanced.

 \begin{table}[t]
\centering

\caption{Performance comparison of confidence evaluation methods with and without refusal instruction(inst).}
\begin{tabular}{l c c c c}
\toprule
\textbf{Confidence Eval} & \textbf{Acc.} & \textbf{Hall.} & \textbf{Score} \\
\hline
\multicolumn{4}{l}{\textit{w/o inst}} \\
None         & \textbf{44.78}       & 55.14\%        & -10.36\%       \\
LLM-Based  & 30.71\%       & 18.17\%        & 12.55\%        \\
Entropy-Based  & 42.23\%       & 43.11\%        & -0.88\%        \\
LLM+Entropy-Based       & 29.03\%       & \textbf{15.54\%}        & \textbf{13.49\%}        \\
\hline
\multicolumn{4}{l}{\textit{w/ inst}} \\
None           & \textbf{31.87\%}       & 12.25\%        & 19.62\%        \\
LLM-Based   & 26.40\%       & 10.21\%        & 16.19\%        \\
Entropy-Based   & 30.49\%       & 10.36\%        & \textbf{20.13\% }       \\
LLM+Entropy-Based        & 24.73\%       & \textbf{9.04\%}         & 15.68\%        \\
\bottomrule
\end{tabular}

\label{Table:confidence}
\end{table}

\subsubsection{Impact of query positioning.}

Figure~\ref{fig:figure_5_6}(c) demonstrates the impact of query positioning on prompt effectiveness across multiple evaluation metrics. Among the three strategies, Query-Append consistently yields the best performance, achieving the highest accuracy and lowest hallucination and missing rates. Query-Surround performs comparably, with only a slight drop in score, indicating that adding an additional query token before the context neither significantly helps nor hurts the response quality. In contrast, Query-Prepend performs noticeably worse across all metrics.

This degradation may stem from a lost-in-the-middle effect~\cite{liu2024lost}, where placing the query before a long context causes large language models to lose attention focus, making it harder to maintain awareness of the query during generation. These findings suggest that positioning the query after the context thereby allowing the model to read and absorb all relevant information before formulating an answer substantially improves response relevance and factuality. For tasks involving lengthy retrieved contexts, adopting a post-context query position is especially important to mitigate attention drift and enhance model robustness.

\subsubsection{Analysis of Confidence Strategies in PruningRAG}
We evaluate PruningRAG with different confidence evaluation strategies and prompt settings to mitigate hallucination, including context sufficiency checks, entropy-based uncertainty estimation, and their combination. Two prompts are compared: a standard prompt and one instructing the model to answer “I don’t know” when unsure. As shown in Table~\ref{Table:confidence}, explicit refusal prompts consistently reduce hallucination rates, while entropy-based evaluation achieves the best accuracy–hallucination balance, especially when combined with explicit prompts. Combining both strategies yields highly cautious responses, suitable for high-stakes applications.

\begin{figure}[t]
	\centering
	\includegraphics[width=1\linewidth]{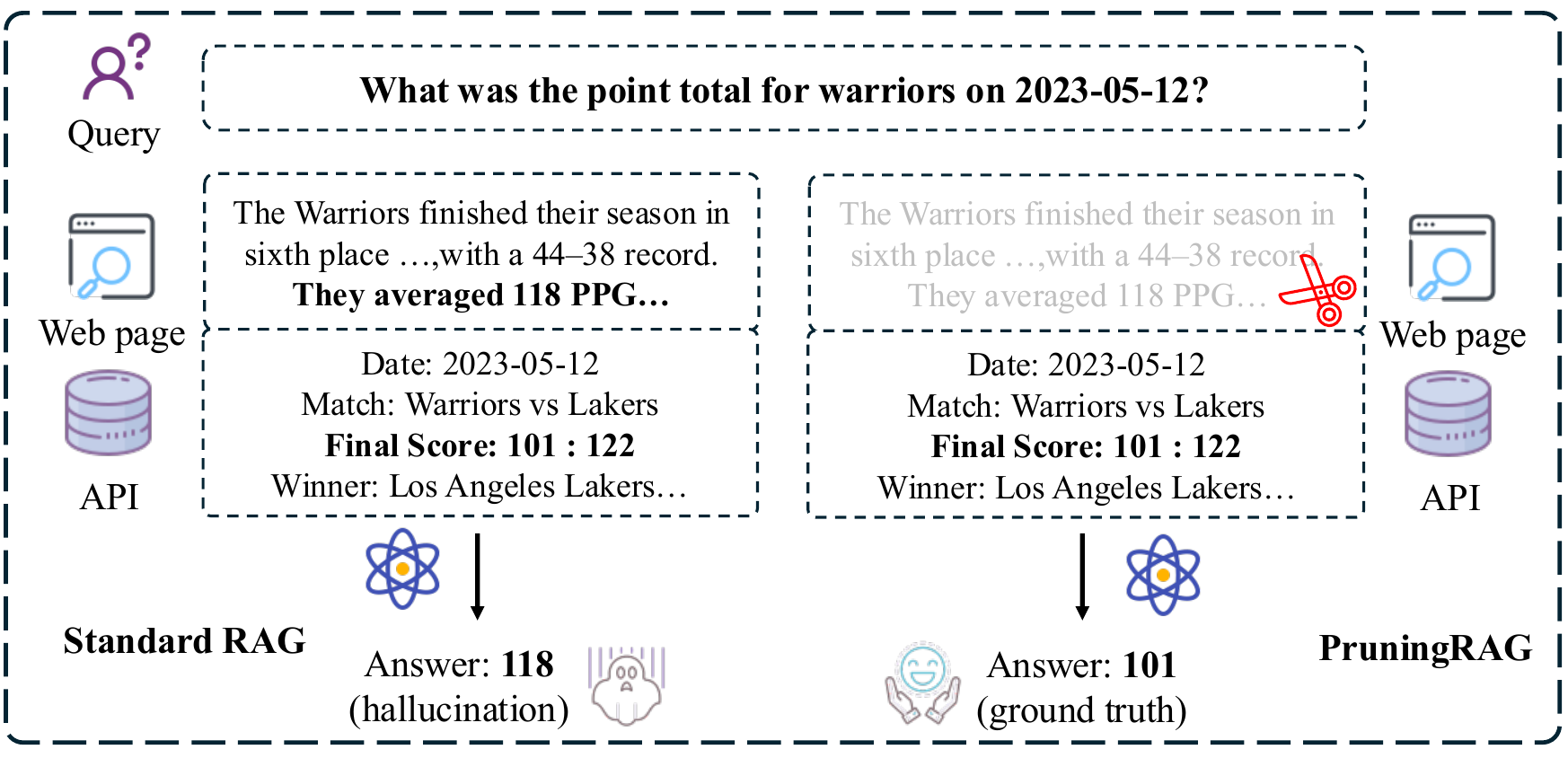}

\caption{
\textbf{Case study of PruningRAG.}
}
	\label{fig:case}

\end{figure}

\subsection{Hyperparameter Sensitivity Analysis }

This section analyzes the impact of retrieval hyperparameters such as chunk size, overlap, and the number of retrieved chunks on the performance of PruningRAG framework.

\paragraph{Impact of   Chunk Size}

Figure~\ref{fig:figure_7_9} (a) demonstrates how chunk size impacts the performance of PruningRAG. Increasing chunk size from 50 to 500 enhances accuracy by providing richer context , but slightly raises hallucination rates as LLM must filter knowledge from more noise. When the chunk size reaches 1000, accuracy declines as the excessive volume of information dilutes relevance making it challenging for the LLM to identify key information. This underscores that a moderate chunk size strikes the optimal balance between context richness and relevance.

\paragraph{Impact of   Chunk Overlap}
As shown in Figure~\ref{fig:figure_7_9} (b), overlap has a limited overall impact on performance. For small chunks (e.g., size 50), it provides minimal benefit due to the narrow context window. For larger chunks (e.g., size 200), overlap slightly improves continuity and accuracy, but may also introduce minor increases in hallucination due to redundancy. In summary, optimizing PruningRAG performance requires balancing chunk size and overlap to ensure sufficient context richness and coherence without introducing excess noise or redundancy.

\paragraph{Impact of Number of Retrieved Chunks}

As shown in Figure~\ref{fig:figure_7_9} (c), increasing the number of retrieved chunks initially improves accuracy, which then plateaus and slightly declines. Hallucination rates show a modest rise at first, stabilize, and eventually increase as accuracy drops.
Unlike the trend observed with larger chunk sizes—where excessive content dilutes focus and reduces accuracy—retrieving too many chunks primarily increases hallucination by introducing overly abundant context. While this context may be relevant, it does not necessarily contribute to answering the question, ultimately lowering the overall score.

\section{Case Study}

Figure~\ref{fig:case} presents a case study illustrating \textbf{PruningRAG}’s effectiveness in resolving conflicts between multiple retrieved sources for the query: ``What was the point total for Warriors on 2023-05-12?''. The standard RAG system is misled by a salient but temporally irrelevant statistic (``118 PPG'') from a web page, producing an incorrect answer. In contrast, PruningRAG filters out such generic content, retaining only the date-specific API data, and thus generates the correct answer (``101''), aligned with the ground truth. This demonstrates the importance of removing non-specific knowledge for time-sensitive factual queries.

\section{Conclusion}
This paper presents a standardized multi-source knowledge dataset and proposes the plug-and-play PruningRAG framework, which applies multi-granular pruning to enhance the integration of diverse knowledge sources. The framework consistently improves performance across various existing RAG variants.
Through our framework, we uncover valuable insights, including the impact of different strategies for leveraging multiple knowledge sources.  Furthermore, we have made our dataset, the PruningRAG framework, code, and experimental results publicly available. We hope that our work will inspire further research into advanced knowledge pruning to better tackle the complexities of multi-source knowledge, contributing to the progress of the RAG community.

\section*{Acknowledgments}
This research was supported by grants from the National Natural Science Foundation of China (No. 62502486), the grants of Provincial Natural Science Foundation of Anhui Province (No. 2408085QF193), Kuaishou Research Project, the Fundamental Research Funds for the Central Universities of China (No. WK2150110032).

\section*{GenAI Usage Disclosure}
No generative AI tools were used to generate content, conduct experiments, or analyze results in this work. Generative AI assistance was limited to minor editing—such as spelling, grammar, and punctuation correction—similar to the use of standard writing aids like spell-checkers or grammar tools. Per ACM guidelines, such use does not require detailed disclosure.

\bibliographystyle{ACM-Reference-Format}
\bibliography{sample-base}

\end{document}